\documentclass[10pt,twocolumn,letterpaper]{article}

\usepackage{cvpr}
\usepackage{times}
\usepackage{epsfig}
\usepackage{graphicx}
\usepackage{amsmath}
\usepackage{amssymb}

\usepackage{booktabs}
\usepackage{graphicx}
\usepackage{pdfpages}
\usepackage{caption}
\usepackage{subcaption}
\usepackage{pifont}%
\newcommand{\xmark}{\ding{55}}%

\newcommand{\boldspacepar}[1]{\par\smallskip\noindent\textbf{#1}}
\newcommand{\emphspacepar}[1]{\par\smallskip\noindent\emph{#1}}

\usepackage[stable]{footmisc}

\usepackage[pagebackref=true,breaklinks=true,letterpaper=true,colorlinks,bookmarks=false]{hyperref}

\cvprfinalcopy %

\def\cvprPaperID{10148} %

\ifcvprfinal\fi

\begin{document}

\title{Unsupervised Model Personalization while Preserving Privacy and Scalability: \\An Open Problem  }

\author{
Matthias De Lange\textsuperscript{1}\\Ales Leonardis\textsuperscript{2} \and Xu Jia\textsuperscript{2} \\ Gregory Slabaugh\textsuperscript{2} \and Sarah Parisot\textsuperscript{2,3}\\Tinne Tuytelaars\textsuperscript{1} \\
\and \textsuperscript{1}KU Leuven \quad \textsuperscript{2}Huawei, Noah's Ark Lab  \quad \textsuperscript{3}Mila\\
{\tt\small \{firstname.lastname\}@kuleuven.be} \quad {\tt\small \{firstname.lastname\}@huawei.com}
}

\maketitle

\begin{abstract}
This work investigates the task of unsupervised model personalization, adapted to continually evolving, unlabeled local user images. We consider the practical scenario where a high capacity server interacts with a myriad of resource-limited edge devices, imposing strong requirements on scalability and local data privacy. 
We aim to address this challenge within the continual learning paradigm and provide a novel Dual User-Adaptation framework (DUA) to explore the problem.
This framework flexibly disentangles user-adaptation into model personalization on the server and local data regularization on the user device, with desirable properties regarding scalability and privacy constraints.
First, on the server, we introduce incremental learning of task-specific expert models, subsequently aggregated using a concealed unsupervised user prior. Aggregation avoids retraining, whereas the user prior conceals sensitive raw user data, and grants unsupervised adaptation. 
Second, local user-adaptation incorporates a domain adaptation point of view, adapting regularizing batch normalization parameters to the user data.
We explore various empirical user configurations with different priors in categories and a tenfold of transforms for MIT Indoor Scene recognition, and classify numbers in a combined MNIST and SVHN setup.
Extensive experiments yield promising results for data-driven local adaptation and elicit user priors for server adaptation to depend on the model rather than user data.
Hence, although user-adaptation remains a challenging open problem, the DUA framework formalizes a principled foundation for personalizing both on server and user device, while maintaining privacy and scalability.
\end{abstract}
\thispagestyle{empty}
\section{Introduction}
Data availability and increased hardware efficiency have made neural networks thrive in a wide range of tasks, competing human-level performance in a variety of tasks \cite{lecun2015deep}. 
However, high performing deep neural network models lead to considerable data requirements, with high capacity models trained on large amounts of labeled data.
Additionally, performance could be significantly increased by personalizing models to user-specific data. Nonetheless, user data cannot be shared directly due to rigorous privacy constraints.
This motivates the need to separate supervised model training on the server from local adaptation to a user's unlabeled personal data. Furthermore, the personalized user model performing tasks locally has the additional benefit of alleviated connectivity requirements.

In this work, we explore this challenge in a class incremental learning setting, relying on the assumption that a user's personal data evolves over time. We define a pragmatic distributed setup comprising a central server connected with a large number of user devices. We assume the server to be a high-end machine endowed with extensive storage capacity and computational resources. By contrast, the compact user device has limited resources for both storage and computation. %
In practice, the number of users may be very high, hence imposing the need for scalable user-adaptation.
A naive approach consists of training a new server model from scratch for each user, resulting in a linearly increasing demand for computational resources. Another infeasible route is to locally finetune user-models, as user devices are confined by limited computational capabilities.
This paper explores a more realistic solution, training a single ensemble of models on the server, subsequently aggregated leveraging user-specific priors.
Besides scalability, two additional hurdles must be tackled: the server is precluded access to local user data by pressing privacy requirements, and user data is typically unlabeled, raising the need for unsupervised adaptation.

In order to tackle these rigorous constraints, we introduce a new Dual User-Adaptation framework (DUA) for model personalization using a task incremental setup. In this setting, tasks are defined as clearly delineated batches of independent and identically distributed (i.i.d.) data, enabling network optimization for a task through multiple iterations over its data. The server trains on the sequence of tasks and provides the user either a general or personalized model. This model can be prone to further local adaptation on the user device.
Hence, DUA disentangles user-adaptation in two phases: 1) the server exploits a model adaptation strategy, with model weight importance as a proxy for user data, and 2) the user device directly adapts to local data using domain adaptation tools. 

In more detail, server adaptation relies on two main components for task incremental learning that are well suited to fulfill our constraints for unsupervised and scalable user adaptation. 
First, Incremental Moment Matching (IMM)~\cite{lee2017overcoming} yields a sequence of task-specific models, restraining new task models to reside close to the previously learned model. Averaging these models presumes a convex-like search space of the loss function, aiming for a single merged model optimal for all tasks. Although weighted averaging provides scalability in the number of tasks, IMM presumes fully supervised model weighting. Therefore, to overcome this need for supervised user data, we incorporate Memory Aware Synapses (MAS)~\cite{aljundi2018memory} deriving parameter importance from unlabeled data, to attain scalable and fully unsupervised \emph{Remote Adaptive Continual Learning} (RACL).

Secondly, DUA establishes further local user adaptation of the model obtained from the server.
Our domain adaptative approach considers the fact that
user and server data distributions resemble two different domains, where we want to transfer domain knowledge from server to user domain. 
A particularly suitable domain adaptation approach is Adaptive Batch Normalization (AdaBN) \cite{li2016revisiting}, which simply collects Batch Normalization (BN) statistics from the target user data. 
Domain knowledge of the user is assumed to reside in these BN statistics, which can be retrieved at low computational cost and without any supervision.
Consequently, this unsupervised setup can enhance any method to become (more) user-adaptive.

The scope of this paper includes user adaptation within the continual learning paradigm, with threefold contributions in this unexplored setup: 
\begin{itemize}
    \item We establish an inherently scalable and privacy-preserving  Dual User-Adaptation framework (DUA), flexibly disentangling user adaptation from the server and local user device.
    \item We introduce a novel benchmark, specifically designed for the evaluation of locally adaptive models in an incremental learning setting. 
    \item We provide empirical evidence supporting the validity of IMM mode-merging with unsupervised MAS importance weights, and find importance weights to depend on the model rather than data.
\end{itemize} 

Adapting to user data with RACL culminates in several advantages for both server and user. 
The server establishes a single sequence of $N$ task-specific models using IMM. As a consequence, independence is imposed on the typically excessive amount of users $L$.
Further, the continual learning setup enables the server to accumulate its knowledge with new arriving batches of data. Hence, this evades rebuilding the server knowledge base from scratch, which would impose time-consuming retraining and storage of all seen data. 
Moreover, storing one model per task instead of its training dataset results in enhanced storage requirements, especially when task data greatly exceeds model size.
Additionally, users only share model parameter importance instead of their raw data, confining shared information to model-specific gradients. %
On top of that, users are not required to label local data as importance is measured in an unsupervised manner. However, when a subset of labeled user data is available, performance can be increased further by local user adaptation, as we will show later.

\section{Related Work}
The DUA framework introduces a new paradigm for user adaptation on the server, resembling federated learning~\cite{mcmahan2016communication}, although completely overturning the purpose.
Federated learning updates a common server model with an aggregated gradient from a distributed database, wherein each user constitutes a node providing local gradients. 
Similarly, DUA solely uses user-specific gradients to acquire better models, but attains decentralized user-personalized models, instead of a general trend-following model. 
Our framework invigorates profound overall user privacy, ensuring no sensible raw user data has to be shared, and additionally tackles the challenging issue of scalability for millions of personalized neural networks. 

Further, sequentially learning multiple tasks by finetuning a neural network results in significant loss of previously acquired knowledge.
Literature on continual learning largely addresses coping with this catastrophic forgetting \cite{de2019continual, parisi2019continual}. Nonetheless, recent works mainly focus on supervised data, leaving the richness of available unsupervised user data unused. 
Following \cite{de2019continual}, these methods can be subdivided into three main categories.
First, \emph{parameter-isolation} methods preserve task knowledge by obtaining task-specific masks \cite{Mallya2017, Mallya2018, Serra2018}, or dynamically extending the architecture \cite{rusu2016progressive}.
\emph{Replay} methods preserve a subset of representative samples of the previous tasks, replayed during training of new tasks. These exemplars can be raw images \cite{lopez2017gradient, chaudhry2018efficient, Rebuffi2017}, or virtual samples retrieved from task-specific generative models \cite{shin2017continual}. Rao \etal~\cite{rao2019continual} extend virtual replay to a completely unsupervised setting based on variational autoencoders. However, this would require exhaustive training on the low capacity edge-device of the user, with only a limited set of available user data, hence infeasible for user personalization.
Finally, \emph{regularization-based} methods impose a prior in the loss function when training the new task.
Learning without forgetting (LwF) \cite{li2016learning} minimizes a KL divergence prior to remain close to the new sample's output on the previous task model, hence distilling previous task knowledge \cite{hinton2015distilling}. Further work \cite{Triki2017} extends this idea with task-specific autoencoders, additionally penalizing new task features to drift away from features deemed important for previous tasks.
Elastic Weight Consolidation (EWC)~\cite{kirkpatrick2017overcoming} introduces a prior on previous-task parameters in a sequential Bayesian framework, Laplace approximated by a Gaussian with diagonally assumed Fisher information matrix (FIM) as precision. 
As the FIM is estimated in the task optimum, Zenke \etal~\cite{zenke2017continual} propose an online approach to estimate precision during training instead.
Furthermore, the FIM relies on the loss gradient $\nabla \mathcal{L}$, whereas MAS~\cite{aljundi2018memory} sidesteps this supervised loss dependency by relying on the output gradient $\nabla F$ instead. 
IMM~\cite{lee2017overcoming} differs from previously discussed methods in first preserving trained task models, which are subsequently merged using FIM importance weights or by averaging.  

For server user-adaptation in our DUA framework, multiple task-specific models are compressed into a single model. 
This is in the same vein to several other works.
Chou \etal~\cite{chou2018unifying} merge two task-specific networks, subsequently finetuned with both task data. Although reduced training time is targeted, it remains unfit for scalable personalization, requiring raw user data and data of all tasks.
Cheung \etal~\cite{cheung2019superposition} superpose different models into a single one from which task-specific parameters can be retrieved. However, adapting models to users linearly increases training time. 
Another compression route distills knowledge~\cite{hinton2015distilling} from teacher networks into smaller nets.
Nevertheless, the focus of this work is model compression to achieve a smaller model for deployment, without addressing scalability to employ user personalization. 

Finally, deep domain adaptation introduces several unsupervised back-propagation based techniques \cite{ganin2014unsupervised,long2016unsupervised}, with state-of-the-art introducing an adversarial loss during training \cite{tzeng2017adversarial,hoffman2017cycada}.
The unsupervised setting befits these methods for adaptation to unlabeled user data.
However, user data is required during training, and therefore unscalable as each personalized model would require training from scratch.

\section{Methodology}
\begin{figure*}[ht]
\centering
\caption{\label{fig:setup}
The Dual User-Adaptation framework (DUA): (1) server user-adaptation involves adaptation to local user data $d_l$ with $\psi$ for each model in $\mathcal{M}$. Aggregating function $\chi$ incorporates all models $\mathcal{M}$ and resulting user priors $\Psi_l$ into single model $\hat{M}_l$.
(2) Local user-adaptation consists of adaptation function $\phi$ mapping $\hat{M}_l$ to the final personalized model $M^*_l$.
}
\includegraphics[clip,trim={0.3cm 0cm 1.6cm 0.3cm},width=0.9\linewidth]{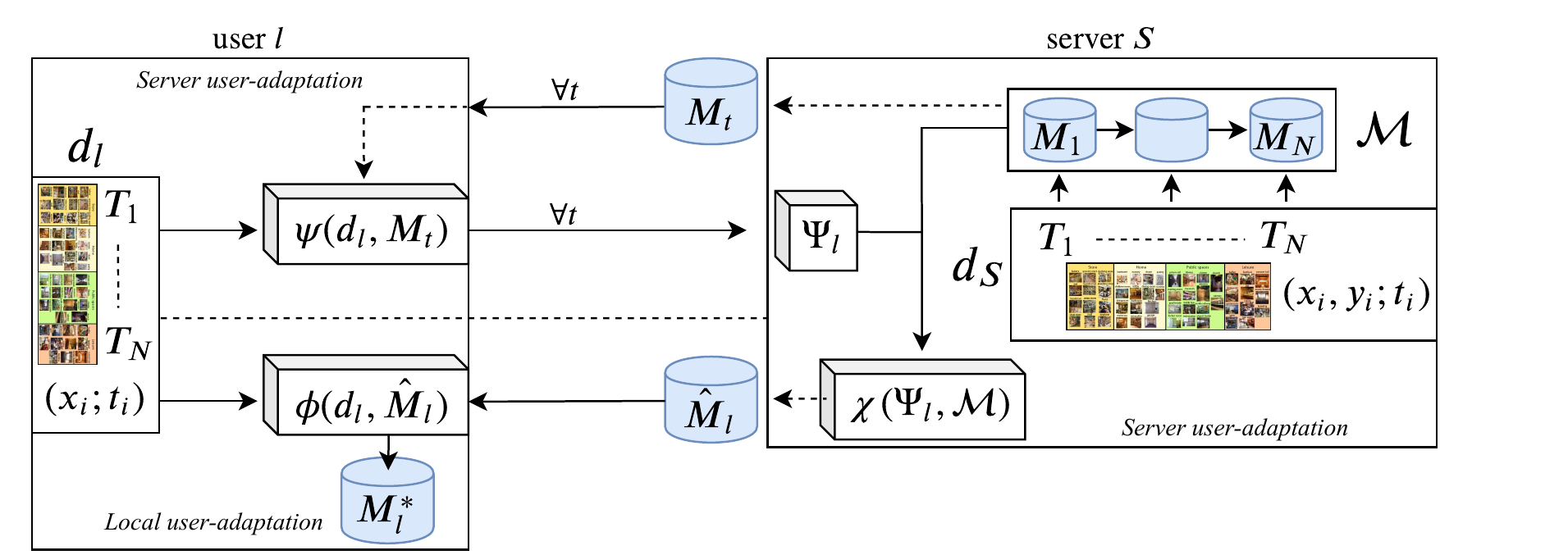}
\end{figure*}%

\subsection{Dual User-Adaptation Framework}
A flexible framework should enable user-personalization both on the server and locally on the user edge-device. For this purpose, the novel Dual User-Adaptation framework (DUA) respectively divides user-personalization in two adaptation functions $\psi$ and $\phi$.
The key to optimal preservation of user privacy is that no raw user data is transmitted.
To achieve this in practice, interactions between user and server are typically encrypted. However, when encryption fails, our framework provides additional concealment of explicit user-data with $\psi$.
Further, adaptation on the server with $\psi$ should be scalable, as training personalized models from scratch would require tremendous resource time for the vast amount of interacting users.
Ideally, both adaptation functions $\psi$ and $\phi$ favor unsupervised local adaptation to limit required user interactions.
The DUA framework is surveyed in Figure~\ref{fig:setup}, with its two user-adaptation phases elaborately discussed in the following.

\boldspacepar{Server user-adaptation.}
Server $S$ has a set of task-specific expert models $\mathcal{M}=\left\{M_1,\dots,M_N\right\}$, trained sequentially on a sequence of $N$ tasks from its  typically labeled data $d_S$. 
The Markov assumption holds for $\mathcal{M}$, with each model depending only on current task data and previous task model, as parameters $\theta_{t+1}$ of model $M_{t+1}$ are initialized with $\theta_{t}$.
$\mathcal{M}$ is continuously extendable with new task models $M_{N+1}$. 

$L$ users interact with $S$, for which a user $l$ provides for each $M_t$ an obscured implicit prior $\psi(d_l, M_t)$ to the server, based on the local raw user-specific data $d_l$. 
To prevent transmission of raw user data, the $\psi$ function uses a task-specific model $M_t$ and corresponding local data $d_l$ to extract concealed user-specific information. This ensures additional privacy safety, as the conveyed user prior is expressed by an implicit proxy, rather than the user data itself.
Ideally, $\psi$ does this without local supervision of the user, as we discuss in Section~\ref{unsupervised}.

Once the server convokes all information in set ${\Psi_l = \left\{ \psi(d_l, M_t) | \  \forall M_t \in \mathcal{M}  \right\}}$ for user $l$, 
aggregating function $\chi$ delivers the ultimate user-personalized model on the server ${\hat{M}_l=\chi\left( \Psi_l, \mathcal{M} \right)}$.

\boldspacepar{Local user-adaptation.}
User $l$ receives $\hat{M}_l$ from the server on the edge device, which can be prone to further local adaptation function $\phi$ to accomplish the final model ${M^*_l = \phi( d_l,\hat{M}_l )}$.
As no data needs to be transmitted, the user can fully exploit its raw local data $d_l$. However, local adaptation is limited in resources, inhibiting exhaustive training procedures for the model.
Flexibility in this framework facilitates $\phi$ to further process an already personalized $\hat{M}_l$ as in the first DUA phase, or any general model without serverside personalization. This endows DUA to extend any method delivering a single model, with adaptation to the local user data $d_l$ as in Section~\ref{method:unsuperlocaldomadap}.

\subsection{Unsupervised Server User-adaptation}
\label{unsupervised}
To enable the server to perform user-adaptation in the DUA framework, we first have to define how to constitute task experts in $\mathcal{M}$, the user-adaptation function $\psi$, and the aggregating function $\chi$ to establish the final personalized model $\hat{M}_l$.
This section explores opportunities in the challenging setup of task incremental continual learning, with only data of the new task $T_n$ available, and inhibiting data access for previously learned tasks.
This is particularly suited for the server, which can learn a new task, discard the new task data $D_n$, and keep only the model $M_n$ to enable incremental learning for further tasks. Thus, server data $d_S$ at any point comprises only new task data distribution $D_n$, from which $\left(x_i, y_i; t_n \right)$ is sampled, respectively constituting the image $x_i$, label $y_i$, and task index $t_n$.

The task experts in $\mathcal{M}$ and the aggregating function $\chi$ are defined following Lee \etal~\cite{lee2017overcoming} with Incremental Moment Matching (IMM). 
Parameter uncertainty is introduced using the Bayesian framework, wherein incremental training of tasks results in a new task posterior after training.
Task posteriors are presumed Gaussian, with the task sequence posterior aggregating these components in a mixture of Gaussians.  
Mode-IMM~\cite{lee2017overcoming} Laplace approximates the mixture with a single Gaussian, and for this assumption to hold, a smooth and convex loss search space is required between the posterior means of the mixture components. Therefore, we adopt the weight and L2 transfer techniques proposed in \cite{lee2017overcoming}, respectively initializing the network with previous task weights, and urging the new task optimum to remain close to the previous task optimum by L2-regularization.
The aggregating function $\chi$ constitutes the mode of the final Laplace approximation of the Gaussian mixture with $N$ components, and is formalized by its mean
\begin{align}
    \hat{\theta}_{l} &=  \frac{1}{\hat{\Omega}_l} \sum\nolimits_{t}^{N}\alpha_t \Omega_t  \theta_t \,, \label{eq_modeIMM_mean}
\end{align}
and precision
\begin{align}
   \hat{\Omega}_l &= \sum\nolimits_{t}^{N}\alpha_t \Omega_{t} \,.\label{eq_modeIMM_IW}
\end{align}
for user $l$, with precision $\Omega_{t}$ of task $T_t$. 
Mixing ratio $\alpha_t$ weighs importance of task $T_t$, subject to $\sum\nolimits_t^N\alpha_t = 1$. 
The balanced tasks in our experiments are deemed equally important.

Further, the user adaptation function $\psi$ should produce an unsupervised implicit user prior from both the raw user data $d_l$ and a given model. Precision indicates the degree of parameter certainty, and therefore resembles a parameter importance measure. As this resembles an implicit prior, we define $\psi(d_l, M_t)=\Omega_t$.
Nonetheless, to estimate task precision $\Omega_{t}$, mode-IMM employs the Fisher information matrix (FIM) similar to \cite{kirkpatrick2017overcoming}.
 The FIM is constituted by 
 second-order derivatives
 of the loss function, thus requiring labeled data.
In contrast, to enable unsupervised importance measures, we exert MAS importance weights~\cite{aljundi2018memory} based on the expected gradient of the L2-norm of the output function
with respect to parameter $\theta^{k}_t$,
\begin{equation} \label{eq_MAS_IW}
    \Omega^k_t = \mathbb{E}_{x\sim D_t} [ \frac{ \nabla \left\lVert F(x;\theta)\right\rVert_2^2}{\delta \theta^{k}_t}]\,,
\end{equation}
with $D_t \in d_l$ the unlabeled user data distribution for which importance is measured. 
Aggregating mode-IMM and MAS importance weights constitutes \emph{Remote Adaptive Continual Learning} (RACL) for server-side user-adaptation in the DUA framework.

\subsection{Unsupervised Local Domain Adaptation}
\label{method:unsuperlocaldomadap}
The second component enabling dual user-adaptation in the DUA framework 
is local adaptation function $\phi( d_l,\hat{M}_l )$, constrained by limited resources on the edge device.
For lightweight adaptation with $\phi$, we could adapt to the user data statistics with Batch Normalization (BN)~\cite{ioffe2015batch}.
During training, each input feature $x_k$ of the BN layer is normalized to $\hat{x}_k$ using current batch statistics.
Subsequently, scaling $\gamma_k$ and shift $\beta_k$ parameters are learned, producing the normalized output $y_k$, with 
\begin{align}
    \hat{x}_k = \dfrac{x_k - \mathbb{E} \left[ x_k \right]}{\sqrt{ \text{Var}\left[ x_k \right]}}  \,,\\
    y_k = \gamma_k \hat{x}_k +  \beta_k \,.
\end{align}
Whereas BN obtains global batch statistics of the training data for inference, Adaptive BN (AdaBN) \cite{li2016revisiting} introduces an unsupervised scheme gathering batch statistics of the target domain data instead.
The main idea is for domain knowledge to reside in the batch statistics, rather than the parameters to optimize. In our setup, the target domain is task-specific user data $d_l$, enabling unsupervised user adaptation with AdaBN.

Relaxing the constraint for unsupervised adaptation, we can assume a labeled subset in the user data. Although this would facilitate finetuning on the user device, the computation and storage limitations both restrain us from computing gradients for all network parameters. Alternatively, we extend AdaBN to this supervised setting (\mbox{AdaBN-S}), additionally training BN layer parameters $\gamma$ and $\beta$ for a few epochs, while freezing all remaining network parameters. This approach significantly reduces the number of trainable parameters compared to finetuning, scaling down computational effort by faster convergence and diminishing storage requirements for the gradients.

\section{User Personalization Benchmarks}
In order to evaluate the DUA framework, we need datasets mimicking user-specific data.
Our experiments comprise three different data setups\footnote{Code available at: \url{https://github.com/mattdl/DUA}}. In all setups, the server data is split into training and validation sets with a ratio of $80 / 20$, and user-data is split into equally sized evaluation and user-validation sets. 
User adaptation techniques such as importance weight estimation or user-specific finetuning solely access the user-validation subset, evading overfitting to the evaluation set or tuning on test data. 

\begin{figure}[]
\caption{TransPrior user transformations following \cite{hendrycks2019benchmarking} with severeness three, including: spatters, elastic transformation, saturation, defocus blur, Gaussian noise, brightness, Gaussian blur, jpeg compression, contrast and impulse noise.}
\centering
        \includegraphics[trim={0.2cm 0.2cm 0.2cm 0.2cm},width=1\linewidth]{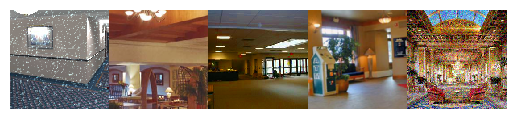} %
        \includegraphics[trim={0.2cm 0.2cm 0.2cm 0.2cm},width=1\linewidth]{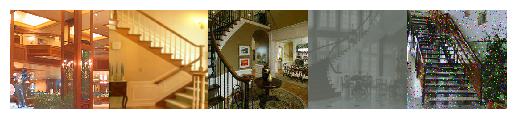} %
\label{fig:transforms}
\end{figure}

Two setups are based on the MIT Indoor Scene recognition dataset (MITIS) \cite{quattoni2009recognizing}, divided into tasks according to the five scene supercategories. Omitting supercategory 'work' as its extra data is too limited, the final task sequence is defined as $\{$home,\, leisure,\, public,\, store$\}$. 
All images have a minimal resolution of $200$ pixels on the smallest axis, are randomly cropped and horizontally flipped during training, and then resized to $224\times224$.
MITIS training data is available to the server, following the continual learning paradigm in only providing access to current task data. %
Evaluation and extra MITIS data are divided over users in the following two schemes:
\begin{enumerate}
\item	\textbf{Category Prior (CatPrior)}. Five users each prefer a random subset of 3 categories per task, acquiring for each preference 250 extra MITIS images. The 20 evaluation images per category are equally divided over users. All user data is mutually exclusive. 
\item	\textbf{Transform Prior (TransPrior)}. Ten users each exhibit a different type of transform following \cite{hendrycks2019benchmarking} with perturbation severeness 3 (in range 1 to 5), permuting 1000 randomly sampled images from extra MITIS data, and all MITIS evaluation data. All user data prior to transformation is identical. See Figure~\ref{fig:transforms} for examples of the ten types of transformed MITIS images.
\end{enumerate}
For all users, Monte Carlo Cross-validation over 5 iterations is performed over the extra MITIS user data, with priors remaining fixed.

The third setup \textbf{(Numbers)} comprises handwritten digits from MNIST \cite{mnist} and Street View House Numbers (SVHN)~\cite{netzer2011reading} data, divided into five tasks of two subsequent numbers as in \{0,1\} to \{8,9\}.
MNIST comprises $28\times28$ images, with $32\times32$ SVHN images center-cropped to match this resolution.
Server data is constituted by both MNIST and SVHN training data, and two users are each characterized by the evaluation data of MNIST and SVHN respectively. Results are averaged over three trained models, initialized with different seeds.

\section{Experiments}

\subsection{Evaluation Setup}
\boldspacepar{Models.}
Experiments with both MITIS setups use AlexNet~\cite{krizhevsky2012imagenet} and VGG11~\cite{simonyan2014very} models, except for the BatchNorm experiments 
which are only applied to VGG11. 
Due to the small input size and simplicity of the Numbers benchmark, a small three-layer MLP with two hidden layers of 100 units suffices for this setup.
For AlexNet and VGG11, we start from models pretrained on ImageNet, while the MLP model is trained from scratch.

\boldspacepar{Evaluation.}
For all experiments, we report average accuracy and forgetting on the final model after training all tasks. Depending on the method, this final post-merging model is either user-specific or the general server model. Results are averaged over all users.

\boldspacepar{Methods} can be subdivided into user-specific and user-agnostic approaches. Table \ref{tab:methodfeatures} summarizes all method features in our user-adaptive setting.

\begin{table}[]
\centering
\caption{Qualitatively comparing features: user-adaptive (Adapt.), unsupervised (Unsup.), scalable (Scal.) and privacy-preserving (Priv.). 
DUA subdivides adaptation on the server ($\psi$) and local user device ($\phi$), with 
MAS importance weights discarding supervision. 
Scalability for user-adaptive methods implies training independent of the number of users $L$.
Shared user-data can be raw $d_l$, gradients of the output function $F(x;\theta)$ or loss $\mathcal{L}(x,y ;\theta)$.
All methods can be extended with unsupervised (AdaBN) and supervised (\mbox{AdaBN-S}) local user adaptation $\phi$.
}
\label{tab:methodfeatures}
\begin{tabular}{@{}lllll@{}}
\toprule
\textbf{Method} & \textbf{Adapt.} & \textbf{Unsup.} & \textbf{Scal.} & \textbf{Priv.}     \\ \midrule
MAS-RACL        & \checkmark ($\psi$)            & \checkmark            & $N$               & $\nabla F$           \\
FIM-RACL        & \checkmark ($\psi$)            &                       & $N$               & $\nabla \mathcal{L}$ \\
Task Experts    & \checkmark            &                       & $N \cdot L$       & $d_l$           \\ \addlinespace
MAS-IMM         &                        & \checkmark            & $N$               &           \\
FIM-IMM         &                        &                       & $N$               &  \\
MAS             &                        & \checkmark            & $N$       &           \\
EWC             &                        &                       & $N$       &  \\
LWF             &                        &                       & $N$       &           \\
Joint           &                        &                       & $N$       &           \\ \midrule

+ AdaBN             &            \checkmark ($\phi$)            &        \checkmark               &        &          \\ %
+ AdaBN-S             &          \checkmark  ($\phi$)            &                       &        &            \\ \bottomrule %

\end{tabular}%
\end{table}

\emphspacepar{User-Specific} methods adapt to the local user-validation set of the user, resulting in a personalized model.
\begin{enumerate}
    \item \textbf{MAS-RACL} is our server user-adaptation method discussed in Section~\ref{unsupervised}, merging task-specific IMM models on the server with unsupervised MAS importance weights obtained from the user-validation set. 
    \item \textbf{FIM-RACL} is a variant of MAS-RACL using the FIM as importance measure to merge the task-specific IMM models. The FIM is constituted by gradients of the loss, hence requiring labeled user data. This baseline serves as a performance reference for the MAS importance weights in MAS-RACL.
    \item \textbf{AdaBN} adapts to the user BN statistics in an unsupervised fashion, only demanding %
    one forward pass
    for all user-validation data (details in Section~\ref{method:unsuperlocaldomadap}).
     \item \textbf{AdaBN-S} collects batch statistics while training BN parameters for a few epochs, hence requiring supervision in the user-validation data (details in Section~\ref{method:unsuperlocaldomadap}).  %
    \item \textbf{Task Experts} are obtained by finetuning each task-specific IMM server model on the raw user validation data for several epochs with low learning rate. This results in an ensemble of task-specific expert networks for each user. %
    This is not scalable and should be regarded an upper bound for user-specific models.
\end{enumerate}{}

\begin{table}[b]
\centering
\caption{Reporting average accuracy (forgetting) for IMM mode-merging with both unsupervised (MAS) and supervised (FIM) importance weights.}
\label{tab:MASvsFIM}
\begin{tabular}{@{}llll@{}}
\toprule
\textbf{Data Setup} & \textbf{Model} & \textbf{MAS-IMM} & \textbf{FIM-IMM} \\ \midrule
CatPrior            & AlexNet        & 67.39 (0.73)     & 67.42 (0.23)      \\
                    & VGG11          & 76.77 (0.30)     & 76.29 (0.43)      \\ \addlinespace
TransPrior     & AlexNet        & 46.51 (-0.14)    & 46.68 (-0.35)     \\
                    & VGG11          & 53.49 (-0.17)    & 53.14 (0.07)      \\\addlinespace
Numbers             & MLP            & 84.36 (-0.40)    & 87.68 (0.07)      \\ \bottomrule
\end{tabular}%
\end{table}%

\emphspacepar{User-Agnostic} methods only access server training and validation data, and therefore are not adapted to the user. 
\begin{enumerate}
    \item \textbf{FIM-IMM} trains a model per task, subsequently merged using a per-parameter importance measure estimated on the server validation data. Identical to mode-IMM in \cite{lee2017overcoming}.
    \item \textbf{MAS-IMM} is a variant of FIM-IMM using MAS importance weights.
    \item \textbf{Joint training} optimizes all tasks at once, accessing all task data simultaneously. This violates the continual learning setup, and is considered as a weak upper bound for performance.
    \item \textbf{EWC} \cite{kirkpatrick2017overcoming} preserves previous task knowledge using FIM-based importance weights. 
    \item \textbf{MAS} \cite{aljundi2018memory} instead uses the gradient of L2 norm of the output function to measure importance.
    \item \textbf{LwF} \cite{li2017learning} uses knowledge distillation with new task data outputs obtained from the previous task network.
\end{enumerate}{}
No gridsearches are performed for forgetting-related hyperparameters as the previous task data is assumed unavailable in the continual learning paradigm. Therefore, the recommended setting of the original works is used. Other hyperparameters are determined from best performance on the Joint baseline, yielding $30$ and $10$ epochs with a learning rate of $1e^{-3}$, and batch size $30$ and $20$ for MITIS and Numbers respectively. After five epochs of unimproved validation accuracy, the learning rate anneals with a factor of $0.1$, stopping early after five more subsequent inferior epochs.

\begin{table*}[!htbp]
\centering
\caption{\emph{Left}: Average accuracy (forgetting) for the three data setups and models, comparing user-specific (RACL) and user-agnostic (IMM) importance weights, both unsupervised (MAS-) and supervised (FIM-). RACL outperforming the corresponding IMM variant is indicated in bold.
\emph{Right}: Qualitatively comparing features user-adaptive (Adapt.), unsupervised (Unsup.), scalable (Scal.) and privacy-preserving (Priv.).}
\label{tab:IMMvsRACL}
\resizebox{\textwidth}{!}{%
\begin{tabular}{@{}llllllllll@{}}
\toprule
\textbf{Method} & \multicolumn{2}{c}{\textbf{Alexnet}}               & \multicolumn{2}{c}{\textbf{VGG11}}                 &  \multicolumn{1}{c}{\textbf{MLP}}     & \textbf{Adapt.} & \textbf{Unsup.} & \textbf{Scal.} & \textbf{Priv.} \\\cmidrule(lr){2-3} \cmidrule(lr){4-5} \cmidrule(lr){6-6}
                & \multicolumn{1}{c}{\textit{CatPrior}} & \multicolumn{1}{c}{\textit{TransPrior}} & \multicolumn{1}{c}{\textit{CatPrior}} & \multicolumn{1}{c}{\textit{TransPrior}} & \multicolumn{1}{c}{\textit{Numbers}} &                              &                             &                         &                                   \\  \midrule
MAS-RACL        & 66.97 (0.88)            & \textbf{47.04 (-0.27)}   & \textbf{77.32 (0.77)}   & \textbf{53.59 (-0.14)}   & 84.01 (-0.22)         & \checkmark                   & \checkmark                  & \checkmark              & \checkmark                        \\
MAS-IMM         & 67.39 (0.73)            & 46.51 (-0.14)            & 76.77 (0.30)            & 53.49 (-0.17)            & 84.36 (-0.40)         & \xmark                       & \checkmark                  & \checkmark              & \checkmark                        \\ \addlinespace
FIM-RACL        & 67.20 (0.73)            & \textbf{47.32 (-0.51)}            & \textbf{76.53 (0.68)}   & \textbf{53.73 (-0.13)}   & \textbf{87.83 (0.30)} & \checkmark                   & \xmark                      & \checkmark              & \checkmark                        \\
FIM-IMM         & 67.42 (0.23)            & 46.68 (-0.35)            & 76.29 (0.43)            & 53.14 (0.07)             & 87.68 (0.07)          & \xmark                       & \xmark                      & \checkmark              & \checkmark                        \\ \bottomrule
\end{tabular}%
}
\end{table*}

\subsection{Unsupervised Moment Matching}
The first experiment studies similarity in IMM mode-merging performance of the original supervised FIM, and the proposed unsupervised MAS importance weights.
The results in Table \ref{tab:MASvsFIM} show similar performance for MAS-IMM and FIM-IMM.
However, we observe a more salient discrepancy for the Numbers setup, where FIM attains $3.32\%$ higher average accuracy, despite $0.47\%$ increased forgetting.
Analyzing importance weights, the first Numbers task \{0,1\} results are an order of magnitude higher compared to subsequent tasks in the sequence. By contrast, FIM importance weights obtain the same order of magnitude over all tasks.
As a consequence, MAS importance outweighs stability in first task knowledge, but deteriorating adaptation to new tasks. 
Further, the Numbers MLP model with few parameters is trained from scratch, with the first task learning only a limited set of discriminating features. The output function magnitude depends only on the features learned for this binary task, and especially the limited MLP network size implies greater output function sensitivity to changes in the intermediate features, which may be substantial when learning the second task.
In contrast, the AlexNet and VGG networks initialize the network with an output function depending on vast Imagenet pretraining.
In conclusion, the MAS importance measure provides competitive outcomes, especially for the setups with pretrained networks.

\subsection{Locally Adapting to the User}
As the MAS importance weights empirically prove a valid alternative precision measure for mode-merging in IMM, we can now adapt to local user data $d_l$ in an unsupervised fashion.
Table~\ref{tab:IMMvsRACL} reports average accuracy over all users for both user-specific and user-agnostic importance weights. The majority of locally estimated importance weights of RACL results in small improvements. 
The benefit of locally adapting to the user seems minimal.

\begin{figure*}[ht]
\caption{Visualizing user importance weight correlation $\rho$ in the CatPrior setup, for the first two tasks. Blue and orange represent weight and bias importance weights respectively. (a) and (b) each compare importance weights of different task data on the same task-specific model, whereas (c) and (d) each use the same data on the two task-specific models.}
\centering
    \begin{subfigure}{0.245\linewidth}
      \centering
            \captionsetup{skip=0.2pt} %
            \caption{$M_1 - D_1 \text{vs}\, D_2$\\ $ \rho=0.82$}%
        \includegraphics[width=1\textwidth]{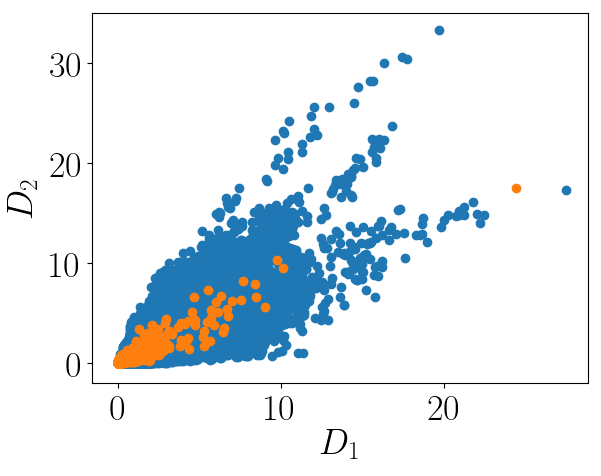} %
    \end{subfigure}%
        \begin{subfigure}{0.245\linewidth}
      \centering
                  \captionsetup{skip=0.2pt} %
      \caption{$M_2 - D_1 \text{vs}\, D_2$ \\ $ \rho=0.73$}

        \includegraphics[width=1\textwidth]{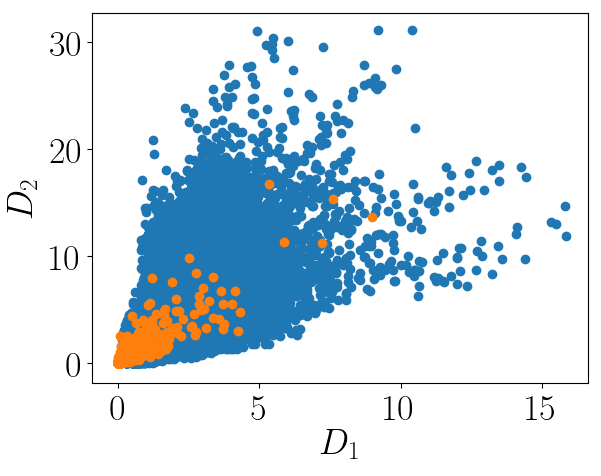} %
    \end{subfigure}%
        \begin{subfigure}{0.245\linewidth}
      \centering
                        \captionsetup{skip=0.2pt} %
      \caption{$M_1 \text{vs}\, M_2 - D_1$ \\ $ \rho=0.55$}

        \includegraphics[width=1\textwidth]{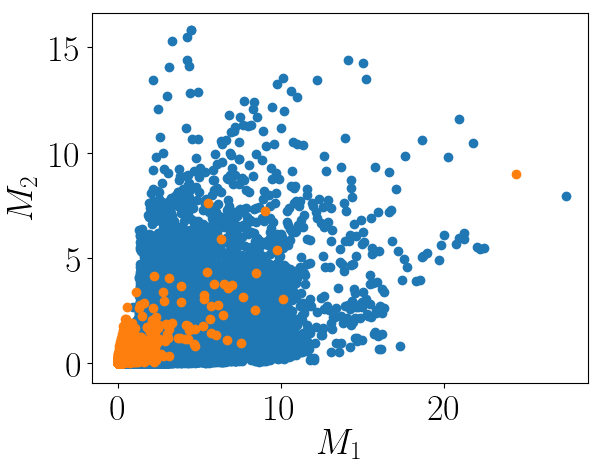} %
    \end{subfigure}%
    \begin{subfigure}{0.245\linewidth}
      \centering
                              \captionsetup{skip=0.2pt} %
      \caption{$M_1 \text{vs}\, M_2 - D_2$ \\ $ \rho=0.58$}

        \includegraphics[width=1\textwidth]{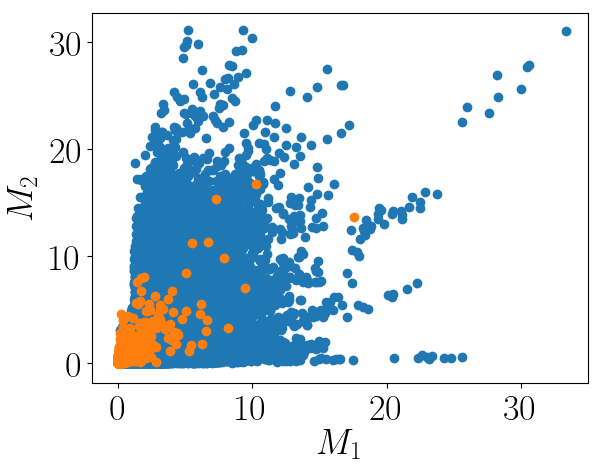} %
    \end{subfigure}

\label{fig:corr}

\end{figure*}

To better understand why this is the case, we further scrutinize data dependency of importance weights measuring Pearson correlation $\rho$.
In Figure~\ref{fig:corr} we consider user-validation data of two CatPrior tasks 'home' ($D_1$) and 'leisure' ($D_2$). Both tasks have corresponding optimized server models $M_1$ and $M_2$, for which our original task incremental setup calculates the importance weights.
In contrast, for this analysis we  compare importance weight correlation for both datasets on the same model, resulting in correlation coefficients $0.82$ and $0.73$ for $M_1$ and $M_2$ respectively (see Figure~\ref{fig:corr} (a) and (b)). This high correlation implies limited dependence of the importance weights on the data.
Next, we compare correlation for the same dataset, yet calculated on the two different models. Correlations for $D_1$ and $D_2$ yield $0.55$ and $0.58$, which is significantly lower, hence indicating higher dependence on the model (see Table~\ref{fig:corr} (c) and (d)).
In conclusion, importance weights represent parameter importance within the specific model rather than the data they are estimated from. As a result, there is little to be gained by estimating them on user-specific data. 

\begin{table*}%
\centering
\caption{\label{tab:BN} Results in the CatPrior and TransPrior setups with model VGG11-BN,
comparing batch normalization on the server data (BN) with unsupervised (AdaBN) and supervised (\mbox{AdaBN-S}) user-adaptive variants.}

\begin{tabular}{@{}lllllll@{}} 
\toprule
\multicolumn{1}{c}{\textbf{Method}} & \multicolumn{3}{c}{\textbf{CatPrior}}             & \multicolumn{3}{c}{\textbf{TransPrior}}           \\ \cmidrule(lr){2-4} \cmidrule(lr){5-7}
\textit{}       & \textit{BN}   & \textit{AdaBN} & \textit{AdaBN-S} & \textit{BN}   & \textit{AdaBN} & \textit{AdaBN-S} \\ \midrule
MAS-RACL        & 58.05 (2.74)  & 58.30 (2.34)   & 60.68 (2.67)     & 30.14 (2.69)  & 30.19 (2.50)   & 32.82 (3.25)     \\
FIM-RACL        & 59.58 (2.14)  & 59.71 (1.61)   & 62.43 (1.84)     & 32.15 (1.53)  & 32.04 (1.33)   & 34.80 (2.13)     \\
Task Experts    & 80.78 (5.61)  & n/a            & n/a              & 68.22 (11.35) & n/a            & n/a              \\ \addlinespace
MAS-IMM         & 55.55 (2.69)  & 55.89 (2.69)   & 58.87 (2.81)     & 29.36 (2.63)  & 29.15 (2.45)   & 31.73 (3.22)     \\
FIM-IMM         & 61.50 (-0.03) & 61.35 (-0.46)  & 63.99 (-0.16)    & 32.08 (1.32)  & 31.86 (1.21)   & 34.48 (2.05)     \\
MAS             & 65.58 (3.96)  & 64.15 (4.04)   & 67.10 (4.66)     & 37.32 (2.64)  & 35.64 (2.88)   & 40.51 (2.69)     \\
EWC             & 66.20 (2.88)  & 64.03 (3.43)   & 67.54 (3.90)     & 37.16 (2.85)  & 35.44 (3.12)   & 40.05 (3.18)     \\
LWF             & 70.76 (0.73)  & 70.37 (0.43)   & 72.73 (1.03)     & 40.22 (0.43)  & 39.51 (0.12)   & 43.07 (0.52)     \\
Joint           & 75.75 (n/a)   & 72.13 (n/a)    & 76.39 (n/a)      & 46.53 (n/a)   & 41.18 (n/a)    & 48.50 (n/a)      \\ \bottomrule
\end{tabular}%
\end{table*}

\subsection{Adapting to the User Domain}
Borrowing ideas from domain adaptation, we can extend any method to become user-specific (see Table \ref{tab:methodfeatures}).
For this experiment, we employ the VGG11 model from previous experiments interspersed with BN layers after each block of convolutional and ReLU activation layers (VGG11-BN).
Table \ref{tab:BN} shows results for the CatPrior and TransPrior setups. Note that Task Experts inherently adapts BN parameters of the VGG model, as it finetunes to the user data $d_l$. 
In general, the unsupervised AdaBN mainly exhibits limited gain for RACL, and seems ineffective for the remaining continual learning methods. 
In contrast, the supervised variant \mbox{AdaBN-S} consistently outperforms user-agnostic BN, with an average gain of $2.64\%$ and $3.44\%$ accuracy in CatPrior and TransPrior setups respectively.
The discrepancy between AdaBN and \mbox{AdaBN-S} performance discloses unsupervised adaptation to remain a challenging open problem.
Remarkably, Joint training with VGG11-BN performs worse than VGG11 results without BN in Table~\ref{tab:IMMvsRACL}. This seems related to overfitting with an observed $10\%$ increase in discrepancy between training and validation accuracy. 
By normalizing over batch statistics, BN layers alleviate the internal covariance shift in the network. For a network prone to overfitting due to few data, this internal covariance shift might instead introduce regularizing noise in the batch, interfering optimization to overfit the training data.
In this respect, even though the server presents a plausibly overfitted model, adaptation to the user domain remains efficacious with \mbox{AdaBN-S}.
Furthermore, the effects of BN on continual learning methods still urge further elicitation, as it is mainly disregarded in current state-of-the-art \cite{de2019continual, parisi2019continual}.

In conclusion, unsupervised adaptation with AdaBN exhibits cumbersome adaptation to the user domain, although gaining significant improvement with a subset of labeled data in AdaBN-S.

\section{Conclusion}
In this work, we proposed a practical Dual User-Adaptation framework (DUA) to tackle incremental domain adaptation to real-life scenarios with numerous users.
This novel user-adaptation paradigm disentangles personalization to both the server and local user device, and combines desirable user privacy and scalability properties, which remain highly unexplored in literature.
We devised benchmarks to scrutinize both types of user-adaptation.

First, adapting models on the server following RACL incurs these scalability, privacy, and additional supervision properties, yet in practice yielded only marginal improvement over a user-agnostic model, due to gradient-based importance weights being largely data independent.

Second, local user-adaptation with a data regularization approach based on adaptive Batch Normalization (AdaBN), and especially its supervised variant ({AdaBN-S}), seem more promising, leading to systematic 
improvements when taking advantage of labeled user-specific data.

User privacy and experience are of major concern, for which our DUA framework forges a principled foundation for dual user-adaptation, 
aspiring to promote further research in this direction.

\section*{Acknowledgements}
\noindent
The authors would like to thank Huawei for funding this research as part of the HIRP Open project.

{\small

}

\end{document}